\documentclass[sigconf]{acmart}
\usepackage{algorithm,amsmath}
\usepackage{algpseudocode}
\usepackage{multirow}
\usepackage{xcolor}
\usepackage{float}
\usepackage{chngcntr}

\algnewcommand{\Inputs}[1]{%
  \State \textbf{Inputs:}
  \Statex \hspace*{\algorithmicindent}\parbox[t]{.8\linewidth}{\raggedright #1}
}
\algnewcommand{\Initialize}[1]{%
  \State \textbf{Initialize:}
  \Statex \hspace*{\algorithmicindent}\parbox[t]{.8\linewidth}{\raggedright #1}
}

\AtBeginDocument{%
  \providecommand\BibTeX{{%
    \normalfont B\kern-0.5em{\scshape i\kern-0.25em b}\kern-0.8em\TeX}}}

\acmConference[KDD '22]{ACM SIGKDD '22: ACM Knowledge Discovery and Data Mining}{June 03--05, 2022}{UNKNOWN}

\begin{document}

\title{Reinforcement Learning in the Wild: Scalable RL Dispatching Algorithm Deployed in Ridehailing Marketplace}


\author{
Soheil Sadeghi Eshkevari\textsuperscript{1},
Xiaocheng Tang\textsuperscript{1},
Zhiwei Qin\textsuperscript{1},\\
Jinhan Mei\textsuperscript{2},
Cheng Zhang\textsuperscript{2},
Qianying Meng\textsuperscript{2},
Jia Xu\textsuperscript{2}
}
 \affiliation{%
  \institution{\textsuperscript{1}DiDi Labs, California, United States}
  \country{}
 }
 \email{{ssadeghieshkevari, xiaochengtang, qinzhiwei}@didiglobal.com}
 \affiliation{%
  \institution{\textsuperscript{2}Didi Chuxing, Beijing, China}
  \country{}
 }
 \email{{jinhanmei, zhangcheng, mengqianying, jiaxujia}@didiglobal.com}







\renewcommand{\shortauthors}{Sadeghi Eshkevari, et al.}

\begin{abstract}

In this study, a scalable and real-time dispatching algorithm based on reinforcement learning is proposed and for the first time, is deployed in large scale. Current dispatching methods in ridehailing platforms are dominantly based on myopic or rule-based non-myopic approaches. Reinforcement learning enables dispatching policies that are informed of historical data and able to employ the learned information to optimize returns of expected future trajectories. Previous studies in this field yielded promising results, yet have left room for further improvements in terms of performance gain, self-dependency, transferability, and scalable deployment mechanisms. The present study proposes a standalone RL-based dispatching solution that is equipped with multiple novel mechanisms to ensure robust and efficient on-policy learning and inference while being adaptable for full-scale deployment. In particular, a new form of value updating based on temporal difference is proposed that is more adapted to the inherent uncertainty of the problem. For the driver-order assignment problem, a customized utility function is proposed that when tuned based on the statistics of the market, results in remarkable performance improvement and interpretability. In addition, for reducing the risk of cancellation after drivers' assignment, an adaptive graph pruning strategy based on the multi-arm bandit problem is introduced. The method is evaluated using offline simulation with real data and yields notable performance improvement. In addition, the algorithm is deployed online in multiple cities under DiDi's operation for A/B testing and more recently, is launched in one of the major international markets as the primary mode of dispatch. The deployed algorithm shows over 1.3\% improvement in total driver income from A/B testing. In addition, by causal inference analysis, as much as 5.3\% improvement in major performance metrics is detected after full-scale deployment.  

\end{abstract}




\maketitle

\begin{figure}[htp]
    \centering
    \includegraphics[scale=.36]{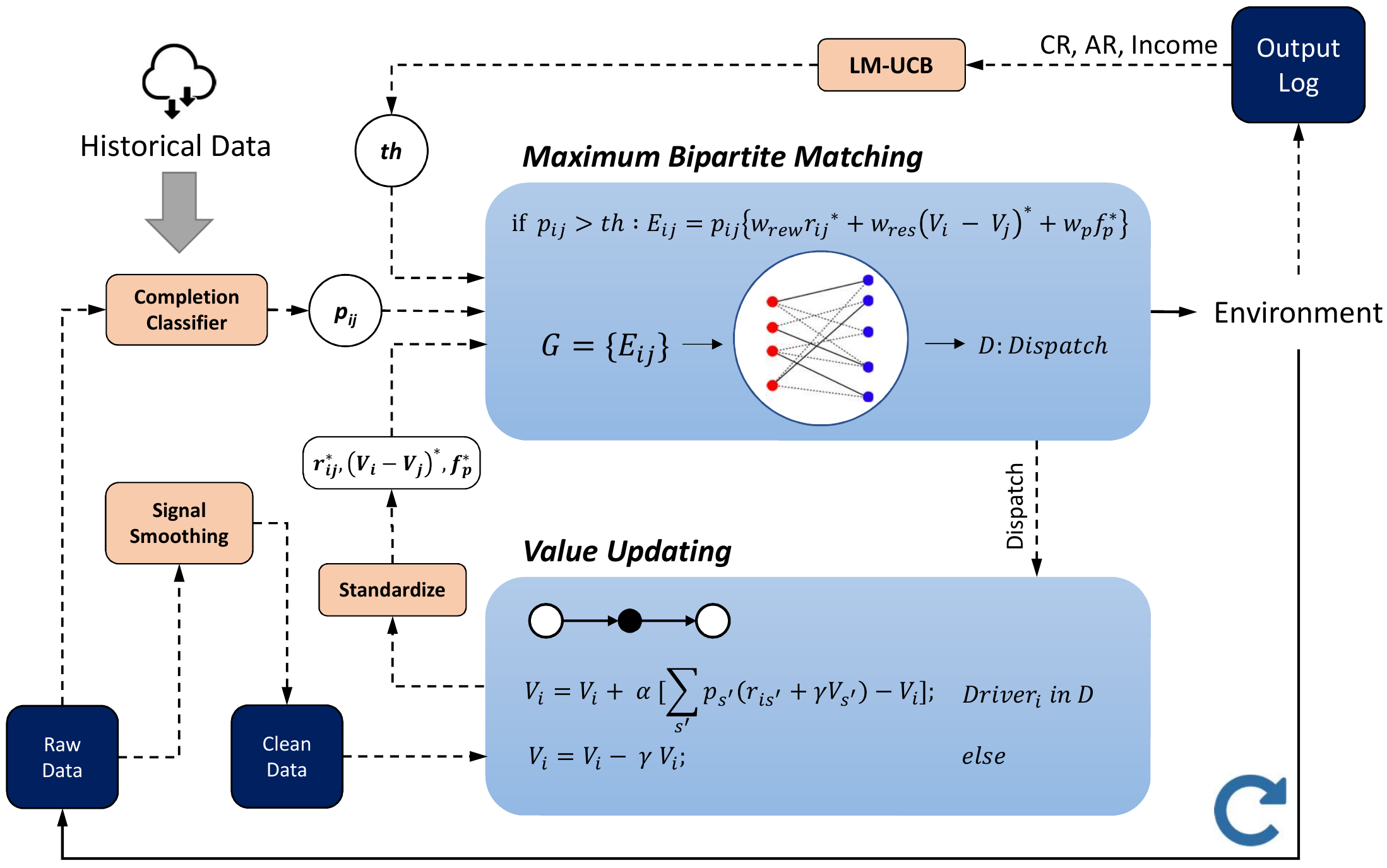}
    \caption{Schematic overview of the algorithm: the real-time dispatching is divided into two main blocks - maximum bipartite matching and value updating. Different rules are employed for each block to increase the efficiency and robustness. A feedback mechanism is also integrated to control the rate of unsuccessful assignments based on global performance metrics. Finally, due to the inherent uncertainties in the assignments, a 
    modified value updating rule is proposed.}
    \label{fig:schematic}
\end{figure}
\setlength{\textfloatsep}{8pt}

\section{Introduction}

Over the past few decades, there has been a growing attention to more personalized modes of transportation due to various promising factors such as flexibility and comfort. In this pursuit, ride-hailing companies such as DiDi, Uber, and Lyft have rapidly provided variety of services to maximize user satisfaction and engagement based on user-specific needs. In these services, one of the primary business objectives is to efficiently connect distributed supply and demand in order to maximize the service quantity while controlling the costs. In addition, targeting flexibility as a principle, such services require to adapt rapidly to significant variations in supply and demand while conforming with a wide range of region-specific variations and regulations. Considering these, the conventional solutions based on empirical findings and previous experiences may not yield a satisfactory service quality. Yet given the access to ubiquitous real-world data, such businesses have a unique opportunity to adapt their technologies and solutions using advanced data-driven and computational methods.

Ride-hailing industry has now turned into one of the major enablers of the mobility on demand (MOD) paradigm by providing a range of transportation information and solutions to users in a fair and competitive marketplace. In order to provide cost-effective solutions for users, such services require to leverage ubiquitous data to improve the efficiencies in stabilizing limited supply (drivers) and asymmetric demand (passengers) spatiotemporally. In addition, the problem is even more complex for ever-growing ridehailing marketplace considering strong nonstationarities in supply and demand regimes, local urban planning factors, and regional regulations. Such factors motivate thorough exploration for unified and transferrable solutions that are based in theoretical grounds and yet are flexibly adapt to real-world deployment challenges. 

The efficiency of a ridehailing business is widely attributed to two major problems: supply dispatch and relocation \cite{sayarshad2017non,yan2020dynamic,JIAO2021103289}. Naturally, there is limited control on demand in the ride-hailing problem since it is mostly dictated by the spatial layouts in urban areas. However, a ride-hailing business can control supply more effectively in different levels and depending on the service. In this study, the emphasis is on the dispatching problem: given a limited supply of drivers and ever-changing demand distribution, as well as rich historical data, how to improve the efficiency of the driver-order assignment algorithm in order to improve business performance metrics such as completion rate (CR) and/or the total driver income. Traditionally, the dispatching problem was handled by a simple real-time search and assignment for an open driver that can pick up the rider in the shortest waiting time (i.e., on-the-fly approach \cite{vazifeh2018addressing} or first-dispatch protocol \cite{yan2020dynamic}). Alternatively, the batching approach collects a collection of open requests and available drivers within a short time interval and then, perform a maximum bipartite matching to assign optimal driver-rider pairs. Conventionally, the edge weights in the bipartite graph are defined as some static quantities such as the price or pickup distance. By using the batching approach with static edge weights, the dispatching problem is broken down into a series of disjoint and myopic assignment problems between supply and demand at any given time. Simulation-based analyses showed that comparing the batching approach with the first-dispatch protocol, the former outperforms in terms of lowering the total waiting time and improving the rate of service \cite{ashlagi2018maximum}. However, the earlier variations of the batching approach do not take into account any long-term considerations when deciding on the drivers' assignment, leading to a spatiotemporal imbalance in supply and demand in a long-run. To address this important concern, non-myopic approaches have been introduced to take into account longer dependencies between rounds of dispatching intervals.


\citet{ozkan2020dynamic} introduced a dynamic matching algorithm that models the arrival of supply and demand by nonhomogeneous Poisson processes that enable to take into account future availabilities when performing the batched matching algorithm. The algorithm shows a consistent improvement when compared with the on-the-fly approach in a simulation-based evaluation and using synthetic data. Yet, the strong reliance on the model-based supply and demand distributions substantially hinders its real-world deployment. The study sheds lights on the fact that a spatiotemporal knowledge regarding the supply and demand distribution can be effective in improving the efficiency of the dispatching algorithm. Alternatively, there has been successful attempts to formulate the dispatching problem as a sequential decision making process with long-term effects. \citet{tang2019deep} formulate the dispatching problem into a Semi Markov decision process (SMDP) and utilize reinforcement learning algorithms to learn the spatiotemporal state values. The proposed algorithm estimates a spatiotemporal state value function for expected return of drivers given the current time and location using an off-policy reinforcement learning paradigm. Given this pretrained function, the edge weights of the matching graph are formulated as a sum of the trip price and the difference in the expected returns between the current driver location and the rider's drop-off location. Using this approach in real-world trials, authors demonstrate notable improvements in the performance metrics including the total driver income and rate of acceptance. To attain spatiotemporal values, the method requires to pass contextual information in addition to the spatiotemporal coordinates to a pretrained neural network. This is found practically challenging when considering the extremely high rate of queries in online deployment. In addition, a generalized value function requires a long list of contextual features as well a sufficiently diverse historical data which increase complexity of training and inference. 

Alternatively, an online state value iteration scheme is recently introduced which is able to update the expected value of the spatial coordinates based on real-time order and dispatch queries rather than offline and massive datasets \cite{tang2021value}. In this framework, the value function is merely dependent on location, however, it continuously evolves based on recent trends in the spatiotemporal demand queries. This technique circumvents the need for a long list of contextual features while still being able to learn non-stationary variations in an online fashion. In addition, maintaining and querying this state value function is practically straightforward which enables a smooth large-scale deployment. Indeed, achieving robustness in such a minimal training framework is challenging; therefore, \cite{tang2021value} suggests a periodic value function resetting based on an offline value function that is trained according to \cite{tang2019deep}. This approach can theoretically address the robustness concern to some extent, however, practical challenges with offline training still persists. In addition, establishing a scalable and effective value mixing scheme is nontrivial. In the present work, we introduce an online state value iteration algorithm that is self-dependent and is equipped with multiple tunable mechanisms that enhance its robustness. In addition, the approach is designed for full-scale deployment and in fact, it is \textit{the first time} such a RL-based online learning method is \textit{deployed and operational} for dispatching in a major international ridehailing market. In the following section, the overall structure of the proposed algorithm is introduced and novel mechanisms are explained. Next, a brief summary of the deployment challenges and solutions are introduced. In Section \ref{sec:experiment}, the experimental campaigns (offline simulation and real-world deployments) and results are presented. 

\vspace{-2mm}
\section{Modified MDP formulation}\label{sec:MDP}
We define parametrized state value functions in the form of $V_s = \theta^T \phi_s$ in which $s$ denotes the state of the MDP and is equivalent to an encoded version of the current location of the agent (driver). $\theta$ is the trainable parameter and $\phi_s$ is a one-hot vector with 1 at coordinate $s$ and zeros elsewhere ($\phi_s^T \phi_{s'} = 0$).  
According to \cite{maei2011gradient}, the complete TD with gradient correction algorithm adapts the following update rule:
\begin{align}
    \theta_s :&= \theta_s + \alpha \delta \theta_s - \alpha \gamma \phi_{s'} (\phi_s^T \omega) \\
    \delta &= r_{ss'} + \gamma \theta^T\phi_{s'} - \theta^T\phi_s
\end{align}

where $r_{ss'}$ is the fare for the driver locating at state $s$ to pick up a rider and drop them off at state $s'$. $\alpha$ is a learning rate and $\delta$ is the gradient in the value estimate based on the new sample $\{s,s',r_{ss'}\} \in \mathbb{D}_t$ where $\mathbb{D}_t$ is the set of driver-order assignments at time $t$. Note that from all drivers participating in a round of matching, a subset might remain idle for which the sample is in $\{s,s,0\}$ form. Considering the orthogonality of the encoding vectors, the update can be simplified as:
\begin{align}
    \theta_s :&= \theta_s + \alpha \delta \theta_s \label{eq:orig_update1}\\  
    V_s :&= V_s + \alpha [r_{ss'} + \gamma V_{s'} - V_s] \label{eq:orig_update2}
\end{align}

Note that in this framework, $\theta_s = V_s$ is the $s^{th}$ entry of the vector $\theta$; the dual notation is adapted to be consistent with both literature of optimization algorithms as well as the classical RL studies. 

\subsection{Expected Update}

In dispatching problem, Equation \ref{eq:orig_update2} is a valid update rule if the on-policy action guarantees a transition from $s$ to $s'$. However, in practice, when the policy proposes an action, the outcome is uncertain and depends on the acceptance by the driver and the passenger. Therefore, the update should take into account the uncertainty of the outcome for a deterministic action. We modify the update rule to the following form to address this issue:
\begin{align}
    V_s :&= V_s + \alpha [\sum_{s'}p_{s'}(r_{ss'} + \gamma V_{s'}) - V_s] \label{eq:expected_update1} \\
    \sum_{s'}p_{s'}(r_{ss'} + \gamma V_{s'}) &= p_c (r_{ss'} + \gamma V_{s'}) + (1 - p_c) (0 + \gamma V_s) \label{eq:expected_update2}
\end{align}
In Equation \ref{eq:expected_update2}, we calculate a pseudo-sampled value estimate for state $s$ given two possible outcomes for an on-policy action for $s\rightarrow s'$ transition: completion and rejection. The probability of completion $p_c$ is estimated by classification models trained with large amount of historical data. The first term in Equation \ref{eq:expected_update2} captures the update if the trip is served and the reward is collected, and the second term considers the likelihood of cancellation with no reward collected and no state transition. 

One may argue that the original updates (Equation \ref{eq:orig_update2}) can be simply used if the value gradient step takes place after actual agents' transitions. It should be noted that practically, this approach encounters deployment challenges in terms of system infrastructure. In order for enacting a post-transition update scheme, a reliable and low-latency feedback system is required that returns actual outcome of ordered transitions in real-time. Therefore, a pre-transition update scheme is practically more preferred and readily deployable. 

\subsection{Reward Smoothing}

Over the course of training, the state value function is updated based on real-time order requests that are submitted to the platform. Each request is associated with a price that is dynamically determined by other sophisticated algorithms as well as marketing strategies (i.e., dynamic pricing and marketing incentives). Ideally, it is preferred to separate such temporary variations from what the value function actually learns in order to improve robustness. For instance, marketing incentives are usually designated for certain spatiotemporal coordinates and/or granted to a subset of users based on their contextual information. These dependencies shall not affect the information used for training the dispatching value function and therefore, raw price values are considered as noisy signals for the value updating. To reduce this noise, we introduce a reward smoothing mechanism that incorporates momentum to the nominal price value in order to control the extent of variation. The momentum is simply a rolling average of the previous sample prices for a certain grid. This process is implemented by the update rule shown in Equation \ref{eq:rew_smooth}.

\begin{align}
    S[grid_o] = \beta. S[grid_o] + (1-\beta) .price_o  \label{eq:rew_smooth}
\end{align}
where $\beta$ is a momentum coefficient, and $S$ is a collection that keeps track of the grid-based smoothed price values. For each order $o$, $grid_o$ and $price_o$ are respectively the starting grid and price of the order. By following this process, we replace the use of raw price values (immediate rewards) in Equation \ref{eq:expected_update2} with the smoothed price. By this mechanism, the model is able to reduce the variance in the updating signals received by platform. A similar mechanism is recently introduced in \cite{dong2021variance} for controlling the stability of the training process in RL. 

\section{Maximum Matching Problem}\label{sec:matching}
In general, real-time dispatching can be divided into two main alternating stages: (1) state value updating and (2) maximum bipartite matching (see Figure \ref{fig:schematic}). The formulations given in Section \ref{sec:MDP} present the rules according to which the online spatiotemporal value function is updated. In this section, an efficient and novel approach is presented for utilizing the value function for the assignment problem (i.e., stage 2). 

\subsection{Graph Edge Standardization}\label{sec:standardization}
Conventionally, the problem of order-driver assignment is formulated as a maximum bipartite matching in which drivers and orders are graph nodes and possible transitions between these two are weighted according to a certain utility function. In RL-based dispatching solutions, the update rule ($\delta$ in Equation \ref{eq:orig_update1}) is used as the utility function \cite{tang2021value}. Despite its logical sense, this utility function does not tune the relative weights between the components - immediate reward and residual value. In addition, it is common in practice to introduce some penalty terms to the utility function for some business purposes and it is challenging to set a weight factor to these penalty terms in the original function. Alternatively, we introduce a new utility function for graph edge weight calculation that is more flexible and address the mentioned challenges. The proposed function has the following form:
\begin{align}
    E_{ss'} = p_{ss'} \cdot [w_{rew} \cdot r_{ss'}^* + w_{res} \cdot (\gamma V_{s'} - V_s)^* - w_{p} \cdot f(ss')^*]
\end{align}\label{eq:edge}
where $E_{ss'}$ is the graph edge weight for a driver-order pair transitioning from $s$ to $s'$, $p_{ss'}$ is the probability of order completion, $w_{rew}, w_{res}$ and $w_{p}$ are weight factors that can tune relative importance of the immediate reward, residual value, and penalty term, respectively. The asterisk denotes a standardized version of the component. The standardization is held in an exponentially moving average manner. Finally, $f(ss')$ is a placeholder for any business-related penalty terms (e.g., pickup distance, waiting time, etc.). The standardization is held as follows:

\begin{equation}
\text{Stdizer\_update: }
  \begin{cases}
  \label{eq:standardize_update}
    m \leftarrow{} \beta_1 m + (1 - \beta_1) x & \\
    v \leftarrow{} \beta_2 v + (1 - \beta_2) (x - m)^2 &  \\
  \end{cases}
\end{equation} 
\begin{align}
    \text{Standardize: }x^* &= Sigmoid((x - m) / \sqrt{v}) \label{eq:standardize_infer}
\end{align}

where $\beta_1$ and $\beta_2$ are hyperparameters. For any given $x$ - can be reward $r$, residual value $(\gamma V_{s'} - V_s)$, or penalty term $f(ss')$ - the standardization mechanism returns a bounded value $x^* \in [0,1]$. Using this utility function, a graph is constructed and the maximum bipartite matching algorithm determines the edges that collectively maximize the utility of the assignment in the current batch. The main advantage of this approach is that here, the components are homogeneous and probabilistic, allowing for any intuitive combination schemes. . To tune $w_{\{res,rew,p\}}$, one can utilize different hyperparameter optimization methods such as Bayesian Optimization (BO). In addition, depending on the application, these weight factors can be constant or time-variant. In this study, we assume weight factors are linear functions of time described by two boundary values at starting and finishing hours of a day. For more simplification, we assume $w_{res} = (1-w_{rew})$. We use BO for tuning the weight factors using our offline simulation platform. In this optimization scheme, the goal is to maximize a desired objective function (e.g., total driver income) as a function of $\{w_{rew}^s, w_{rew}^f, w_{p}^s, w_{p}^f\}$ in which $s$ and $f$ denote values at starting and finishing hours. 

\subsection{Real-time Adaptive Graph Pruning using Limited-Memory UCB (LM-UCB)}

Once graph edges are calculated using the rules presented in Section \ref{sec:standardization}, a bipartite graph is constructed and a maximum matching algorithm identifies the optimal subset of pairs \cite{kuhn1955hungarian}. One likely outcome is that within some batches of driver-order assignments, the majority of possible matches has low completion probability ($p_c$ in Equation \ref{eq:expected_update2}). In such occasions, if the graph is constructed using all possible edges, the majority of the assignments will be eventually canceled. The drawback of cancellation is twofold: the immediate reward is missed, and the available driver is pointlessly relocating and cannot participate in the next round(s) of assignments. Therefore, it is more efficient if one can prune the driver-order graph based on the probability of the pairs in order to control the impact of cancellation. 

Various approaches can be held for the graph pruning task including filtering edges by a fixed probability threshold. This approach is simple, yet may be inferior when dealing with different supply-demand levels. For instance, during peak hours, the graph complexity is high (assignment options are abundant) and therefore, a higher probability threshold is proper; but that is not the case for quite hours. This implies that a dynamic threshold adjustment is the preferred choice. The dynamism can be defined based on expert knowledge (e.g., a nonlinear function proportional to the temporal demand level) or more flexibly, can be an adaptive mechanism that learns from real-time feedback. In this study, a multi-arm bandit (MAB) solution is proposed for this purpose that takes advantage of the latter idea.

In MAB, at every timestep the agent has to enact an option from multiple choices so that in the long-run, the cumulative return is maximized. In dispatching problem analogy, arms are defined as a set of discrete probability thresholds $TH = \{th_1, th_2, ..., th_N\}$ and the objective is to maximize a utility function consisting of primary business metrics. In practice, three business metrics are monitored for evaluating dispatching algorithms: (1) total driver income, (2) completion rate, i.e., total number of served trips divided by the total number of requests (CR), and (3) answer rate, i.e., total number of accepted trips divided by the total number of requests (AR). As previously mentioned, an order-level and real-time feedback for the online algorithm is practically impossible due to the high rate of required communication. However, these globally defined metrics can be monitored in real-time with less complexity. Given that, a UCB-based bandit solution termed as limited-memory UCB (LM-UCB) is developed that is inspired by the discounted UCB proposed by \citet{kocsis2006discounted} - see more details in the appendix and Algorithm \ref{alg:USB}. 
 
\section{Reinforcement Learning in the Wild}\label{sec:method}


When proposed mechanisms are combined, a full picture of the online reinforcement learning algorithm will be established. The procedural description of the proposed algorithm is presented in Algorithm \ref{alg:pipeline}. In general, the algorithm requires a set of collection instantiation and hyperparameters to initiate the procedure. $V$ is a tabular form of the state value function and is initialized by zero for all spatial coordinates. $S$ is a dictionary of smoothed rewards for spatial coordinates and is initialized by zero. This dictionary is updated by the outstanding requests at every round of dispatch - shown in line \ref{line:rew_smooth} in Algorithm \ref{alg:pipeline} - via Equation \ref{eq:rew_smooth}. $R^*$, $dV^*$, and $P^*$ are tuples that hold standardization parameters $m,v$ for the three components of the edge weight: immediate reward (price), residual value, and penalty term, respectively. In lines \ref{line:std_infer1} to \ref{line:std_infer3} of Algorithm \ref{alg:pipeline}, $r$ (order price), $dv$ and $p$ (order-driver penalty) that are attributes of a single order-driver pair, are standardized using these tuples that are substituted in Equation \ref{eq:standardize_infer}.  In addition, the parameters of the standardization tuples are updated in line \ref{line:std_update1} of Algorithm \ref{alg:pipeline} based on Equation \ref{eq:standardize_update}. $feedback$ is a monitoring mechanism that record statistics of the settled orders in the previous rounds of dispatch. It can be as simple as a log of outputs that register the number of total requests as well as completed and cancelled trips. This log is updated in line \ref{line:feedback_update} of the procedure and is utilized in line \ref{line:feedback_infer} to calculate $CR_t$ and $AR_t$, the real-time performance metrics. Finally, if the current time is divisible by threshold adjusting time interval $C$, the LM-UCB mechanism adjusts the graph pruning threshold $th$ based on $CR_t + 0.1 AR_t$ - line \ref{line:thresh_update}.

Updating the value function using online samples shown in Equation \ref{eq:orig_update1} is analogous to the update rule in stochastic gradient descent (SGD). Therefore, mathematical practices that improve the performance of the vanilla SGD algorithm can be applied in this scenario. In particular, ADAM \cite{kingma2014adam} is a widely used optimization algorithm that can accelerate the learning process in which two well-known techniques - SGD with momentum and Root Mean Square Propagation (RMSP) - are combined. In the present work, we use ADAM for the gradient ascent updates of the state value functions. For this purpose, a dictionary of ADAM parameters with spatial coordinates as the key is defined by $LR$. Intuitively, the integrated ADAM mechanism can adapt to the variance, magnitude and frequency of value updates.


In addition to the described methods and dictionaries, the algorithm requires multiple hyperparameters to be defined. $W_{rew}$ and $W_p$ are sets of hypeparameters that can model arbitrary functions for $w_{rew}$ and $w_p$. In this study, a linear interpolation between two terminal values is used to calculate $w_{rew}$ and $w_p$; therefore,  $W_{rew} = [w_{rew}^s,w_{rew}^f]$ and $W_{p} = [w_{p}^s,w_{p}^f]$ in which $s$ and $f$ denote boundary values - start and finish. As explained in Section \ref{sec:standardization}, one approach to tune these hyperparameters is to use BO. $\gamma$ and $C$ are RL discount factor and time interval for $th$ adjustment. The completion probabilities are required to construct edge weights as well updating value functions. For that, a pretrained classification model - \textit{Cancel\_Model} - is inferred in line \ref{line:cancel_model} based on contextual features of the order-driver pair to estimate $P_C[od]$. This probability is utilized in lines \ref{line:edge} and \ref{line:val_update} for calculating graph edge weights and well as value updates. This complete procedure shown in Algorithm \ref{alg:pipeline} is the basis for the method evaluations in the next sections.
\vspace{-2mm}
\section{Deployment Infrastructure}\label{sec:deployment}

\begin{figure}[h]
    \centering
    \includegraphics[scale=.33]{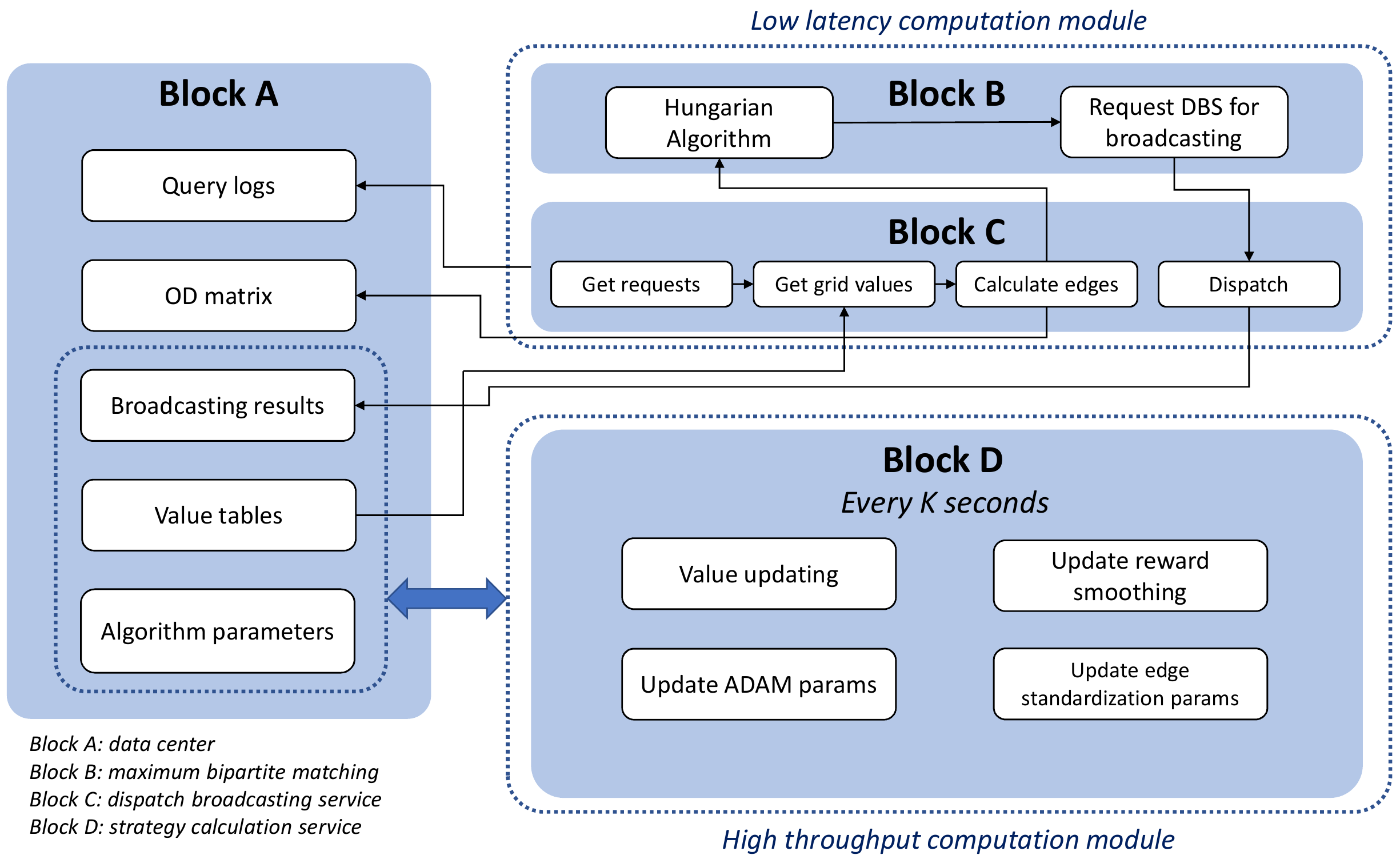}
    \caption{Deployment infrastructure scheme: the backend process is categorized into multiple tasks including data storage (Block A), low-latency analyses (Blocks B-C) and high throughput computations (Block D). All together, this organization has enabled seamlessly performing the RL-based dispatching algorithm in full-scale.}
    \label{fig:deploy}
\end{figure}

The present work is unique in its full deployment in some major international cities through DiDi's marketplace. For this purpose, a backend infrastructure is designated in order to satisfy the high-stake requirements of an online service while seamlessly implementing the proposed RL-based algorithm. Overall, Algorithm \ref{alg:pipeline} is divided into two main segments: (a) order-driver assignment and (b) state value updating. Some updates in the latter segment can potentially be applied within segment (a); for instance, standardization parameters can be updated before inference in line \ref{line:std_infer1}. However, for a large-scale deployment, separating operations within segments (a) and (b) is necessary in order to have low-latency performance. In this framework, fairly inexpensive computations in segment (a) are repeated every two seconds while bulk calculations such as state value, standardization parameters, and reward smoothing updates are held in segment (b) with frequency of once every 10 seconds (defined by $T_{up}$). The deployment infrastructure is also customized to match with this procedure as shown in Figure \ref{fig:deploy}. In this figure, the low-latency and high-throughput computation modules are responsible for the order-driver assignments and algorithmic updates, respectively and similar to the organizations presented in Algorithm \ref{alg:pipeline}. For more more details, see Section \ref{sec:deployment} of the appendix.

\begin{algorithm}[]
\caption{Reinforcement Learning in the Wild (RLW) Procedure}\label{alg:pipeline}
\begin{algorithmic}[1]
\Initialize{$V, S, LR, R^*, dV^*, P^*, feedback , th, Cancel\_Model, T_{up}$}
\Inputs{$W_{rew},W_p,\gamma, C$}
\State $Batch \gets \{\}$
\For{$t = 0,1,2,..,T$} \Comment{$t$ denotes rounds of dispatch}
\State $O_t, D_t, G_t \leftarrow{}$ observation(t)
\State \texttt{// Order-Driver Assignment}
\State $w_{rew} = $ Interpolate$(W_{rew}, t)$
\State $w_{p} = $ Interpolate$(W_{p}, t)$
\State $w_{res} = 1 - w_{rew}$
\State $E, P_C = \{\},\{\}$
\For{$od$ in $G_t$} \Comment{od is an order-driver pair}
\State $o,d =$ order and driver for od
\State $r = $ Standardize$(R^*,S_o)$ \Comment{$S_o$ is smoothed reward}\label{line:std_infer1}
\State $dv = $ Standardize$(dV^*,\gamma V_{o} - V_{d})$\label{line:std_infer2}
\State $p = $ Standardize$(P^*,o,d)$\label{line:std_infer3}
\State $P_C[od] = Cancel\_Model(od)$ \label{line:cancel_model}
\If {$P_C[od] > th$}
\State $E_{od} = P_C[od] \cdot(w_{rew} \cdot r + w_{res}\cdot dv - w_p\cdot p)$ \label{line:edge}
\EndIf
\EndFor
\State $A \leftarrow{}$ Hungarian Algorithm(E) \cite{kuhn1955hungarian} 
\State $Batch.append(A)$ \Comment{$A$: list of batch assignments} 
\State \texttt{// State Value Updating}
\If{$t$ mod $T_{up} == 0$}

\For {$type, item$ in $Batch$}:
\If {$type$ is dispatch}
\State $o,d =$ order and driver for $item$
\State $S \leftarrow{} $ Reward Smoothing$(S, r_o)$ \Comment{Equation \ref{eq:rew_smooth}} \label{line:rew_smooth}
\State $p_c = P_C[od]$
\State $\delta = p_c\cdot (S_o + \gamma V_o - V_d) + (1-p_c)\cdot (\gamma V_d - V_d)$ \label{line:val_update}
\State $R^*\leftarrow{}$ Stdizer\_update$(R^*, S_o)$ \Comment{Equation \ref{eq:standardize_update}} \label{line:std_update1}
\State $dV^*\leftarrow{}$ Stdizer\_update$(dV^*, \gamma V_o - V_d)$ \label{line:std_update2}
\State $P^*\leftarrow{}$ Stdizer\_update$(P^*, o, d)$ \label{line:std_update3}

\ElsIf {$type$ is idle}
\State $\delta = \gamma V_d - V_d$
\EndIf
\State $V_d, LR_d \leftarrow{}$ Adam Update$(LR_d, \delta)$
\EndFor
\EndIf
\State Return driver assignments based on $A$
\State Update $feedback$ based on settled rider requests \label{line:feedback_update}
\State $CR_t, AR_t \leftarrow{feedback}$ \label{line:feedback_infer}
\If {$t$ mod C == 0}
\State $th \leftarrow{} $ LM\_UCB$(CR_t, AR_t, th)$ \Comment{Algorithm \ref{alg:USB}} \label{line:thresh_update}
\EndIf
\EndFor
\end{algorithmic}
\end{algorithm}

\vspace{-2mm}
\section{Experiment}\label{sec:experiment}

Based on the introduced procedure in Algorithm \ref{alg:pipeline}, the new dispatching model is implemented both in offline simulation as well as full-scale deployment, and evaluated against state-of-the-art and state-of-the-practice models. 

\subsection{Offline Simulation with Real-World Data}

For this analysis, we use the internal simulation environment that enables executing different dispatching strategies in various cities and dates using real-world data from DiDi's daily operation. In particular, three major cities from international marketplace are used as the test ground for the offline evaluation. The statistical characteristics presented in Table \ref{tbl:city_stats} for Cities I to III imply distinctive operational patterns. For instance, City I which has the highest population, shows moderate cancel rate and answer rate (AR) with average trip lengths that are $\sim 15\%$ shorter than Cities II and III. In addition, City II with the second highest population shows the highest cancellation rate while having longer trips. City III shows significantly lower AR compared to the other two. It is also the smallest city among targeted domains considered for offline simulation. Note that answer and cancel rates presented in Table \ref{tbl:city_stats} are minmax scaled. In this offline evaluation, multiple models are compared with the proposed algorithm. A brief description of the models is as follows:
\begin{table}[htp]
\centering
\caption{Statistics information for multiple international marketplaces that are targeted for method evaluation. Answer and cancel rates are normalized to values between 0 to 1.}
\label{tbl:city_stats}
\begin{tabular}{lclclclcl} 
\toprule
City & population & answer & cancel & mean trip   \\ [0cm]
 & scale & rate & rate &  length (m)  \\ [0.2cm]
\hline
I    & High       & 0.63         & 0.88     & 5.54e+03          \\
II    & High      & 0.98         & 1.00     & 6.41e+03          \\
III    & Medium   & 0.00         & 0.55     & 6.43e+03          \\
IV    & Medium    & 1.00         & 0.32     & 5.90e+03          \\
V    & Low        & 0.17         & 0.19     & 6.31e+03          \\
VI    & Low       & 0.73         & 0.00     & 4.34e+03          \\
\bottomrule
\end{tabular}
\end{table}
\setlength{\textfloatsep}{8pt}

\noindent \textbf{Baseline:} A myopic method that minimizes total pickup distances between a batch of available drivers and riders using maximum matching. The batched matching is repeated every two seconds.

\noindent \textbf{V1D3:} A predecessor for the present method introduced in \cite{tang2021value} which utilizes an on-policy value iteration algorithm to learn and employ a spatial value function for driver dispatching.  In this method, the inference requires an intermittent value mixing between the online value function and CVNet to accelerate the process of learning. In this study, however, we use the online value function individually as it is more comparable to the the present work. The original V1D3 model \cite{tang2021value} does not incorporate the probability of completion $p_c$ as it is presented in Equation \ref{eq:expected_update2}. In this version of the algorithm, the probability of completion is multiplied by the raw value update for both value updating as well the graph edges for the order-driver assignment. This turned out to be effective in improving the original V1D3 algorithm. 

\noindent \textbf{CVNet-B:} An improved version of deep reinforcement learning model based on off-policy training of a value function approximator proposed in \citet{tang2019deep}. In this method, the value function is trained using historical data and is inferred for online dispatching with no intermittent updates. The approximate value function is modeled by a deep neural network which receives spatiotemporal coordinates to estimate the value. Compared to the original version, the model is improved by reward smoothing and inclusion of completion probability $p_c$ in the edge weight calculation.

\noindent \textbf{RLW:} This is the proposed model based on Algorithm \ref{alg:pipeline} that incorporate probabilistic TD updates, grid-based adaptive learning rates, reward smoothing, edge standardization, and feedback loop based on LM-UCB. In this algorithm, value function is updated after action determination but before the action occurrence, and therefore, the expected update is used. In addition, the learning rates are tuned by ADAM optimizer. The immediate reward in the update calculation is a smoothed version of the trip pricing. Edge standardization allows to control the contribution level of each graph edge component. 
Via standardization, the weight factors associated to the graph edge components are potentially transferrable. The feedback loop mechanism takes advantage of a UCB model to control the rate of cancellation. The UCB objective function is chosen as $CR_t + 0.1 AR_t$ in which $CR_t$ and $AR_t$ are cancellation rate and acceptance rate calculated for the latest threshold tuning interval, e.g., C = 60 seconds. 

Using the offline simulation environment, the performance of the candidate models are evaluated in the three cities over three days. The performance of a dispatching algorithm can be monitored through different lenses; in this study, we focus on three major performance metrics that have high impact in the overall efficiency of the dispatching strategy: (1) completion rate (CR), (2) answer rate (AR), and (3) total driver income which is proportional to the cumulative monetary value of the completed trips. The simulation environment returns these metrics for each city-day-model simulation and the daily-averaged improvements with respect to the baseline model are presented in Table \ref{tbl:offline_res}. 

\begin{table}[]
\centering
\caption{Summary of major performance metrics in comparative analysis using offline simulation: state-of-the-art methods are used as the reference models. The simulation environment uses real-world data from DiDi’s ride-hailing platform in three major cities. Results are averages of three days and the means and variances across days are reported. Values present the percent of improvement with respect to the baseline model.}
\label{tbl:offline_res}
\begin{tabular}{cllll}
\toprule
\multicolumn{1}{l}{City} & Method  & CR (\%)            & AR (\%)            & Income (\%)      \\ \hline
\multirow{5}{*}{City I}  & Baseline& 0.00±0.00  & 0.00±0.00  & 0.00±0.00   \\
                         & V1D3    & 3.37±0.54  & 5.70±1.22  & 4.02±0.51  \\
                         & CVNet-B & 2.52±0.65  & 4.34±0.57  & 2.94±0.51  \\
                         & RLW & \textbf{3.57±0.54}  & \textbf{7.08±0.49}  & \textbf{4.28±0.45}  \\ \hline
\multirow{5}{*}{City II} & Baseline& 0.00±0.00  & 0.00±0.00  & 0.00±0.00  \\
                         & V1D3    & 1.39±0.45  & 2.70±0.32  & \textbf{1.96±0.27}  \\
                         & CVNet-B & 1.05±0.10  & \textbf{8.29±4.67}  & 1.53±0.31  \\
                         & RLW & \textbf{1.64±0.27}  & 3.46±1.44  & 1.36±0.28  \\ \hline
\multirow{5}{*}{City III}& Baseline& 0.00±0.00  & 0.00±0.00  & 0.00±0.00  \\
                         & V1D3    & 4.92±0.80  & 5.20±0.46  & 2.88±0.33  \\
                         & CVNet-B & 5.17±1.45 & 5.21±0.96 & 3.11±0.75 \\
                         & RLW & \textbf{5.48±0.93} & \textbf{6.32±0.98} & \textbf{4.20±0.54} \\ \bottomrule

\end{tabular}
\end{table}


From Table \ref{tbl:offline_res}, in two out of three examined cities, the RLW model outperform others in all three metrics. In terms of AR, RLW is the superior model in City I and City III implying the efficacy of the proposed algorithm in learning the spatiotemporal demand distribution. From Table \ref{tbl:city_stats}, AR is the lowest in City III based on the historical data. This suggests that the collocation between available drivers and the spatiotemporal demand is practically imbalanced. Therefore, an RL-based dispatching algorithm can be highly effective. This is confirmed by the large margin of improvement in all three metrics in City III when following RLW. 

Results in City II are in odds with other cases. While in terms of CR, the RLW model is superior, the total driver income and AR are more boosted by the V1D3 and CVNet-B models, respectively. From statistical characteristics of the cities presented in Table \ref{tbl:city_stats}, City II is a city with highest rate of cancellation and relatively longer requested trips. In addition, the gap between acceptance rate and the cancellation rate is largest among all cases. This implies that while the drivers mostly accept the recommended trips, riders are more likely to cancel the trip after acceptance by the driver. This is detrimental for the performance of a dispatch strategy because within the time frame in which the driver is relocating toward the pickup spot for a trip that will eventually become cancelled, it cannot participate in other dispatching assignments. In such scenarios, CVNet-B that is following a pre-trained value function performs quite efficiently. By learning from large amount of historical data, its value function reflects spatial distribution of high acceptance probability coordinates for any given time of the day. Note that in contrast, the online RL methods such as V1D3 and RLW learn from limited samples in real-time and therefore, the evolving value functions can yield high rate of cancelled assignments in early hours. Comparing RLW and V1D3, the latter is performing better in terms of total income which is perhaps a result of lower residual value and higher price contributions in the un-standardized edge weights. 

\begin{table}[htp]
\centering
\caption{Tuned weight factors from BO: values are used for edge weight calculation based on standardized components.}
\label{tbl:wfs}
\begin{tabular}{lllllll}
\toprule
        & $w_{rew}^s$    & $w_{rew}^f$ & $w_{res}^s$    & $w_{res}^f$    & $w_{p}^s$    & $w_{p}^f$    \\ \hline \rule{0pt}{3ex}   
City I  & 0.430 & 0.008 & 0.570 & 0.992 & 0.002 & 0.004 \\
City II & 0.760 & 0.190 & 0.240 & 0.810 & 0.220 & 0.550 \\
\bottomrule
\end{tabular}
\end{table}

Regarding the RLW model, a preliminary analysis was necessary to tune weight factor hyperparameters for the graph edge weight calculation. For this purpose, we conducted independent BO campaigns for Cities I and II. In general, our analysis resulted in $[w_{rew}^s,w_{rew}^f] \sim [1,0]$ which is strongly intuitive (exact values are presented in Table \ref{tbl:wfs}). This finding suggests that the algorithm prefers to dispatch drivers mostly based on immediate rewards at the early hours in a day when the value function is not sufficiently learned. As the time proceeds, more samples are processed and the value function becomes more reliable, and consequently, $w_{rew}$ declines to put more weight on the $dv$ - see line \ref{line:edge} in Algorithm \ref{alg:pipeline}. In addition, by comparing results from Cities I and II, weight factors for the penalty term (here, proportional to the pickup distance) are significantly higher in the latter case. This is aligned with the previous discussion regarding higher risk of cancellation in this city. By higher values for $w_{p}$, the algorithm intends to avoid assignments based on longer pickup distances that are more likely to cancel. 

\subsection{Large-Scale A/B Testing}

In addition to the offline simulation with real data, the proposed approach is deployed in full scale in multiple international cities through DiDi platform to provide dispatch recommendations for the affiliated drivers. In the first scale of deployment, we conducted statistical A/B testing campaigns in five cities with significantly distinct topological and statistical characteristics (Cities I, II, IV, V, and VI). Due to technical and business-related constraints and challenges, the experiment was held in a two-week time period based on a limited version of the proposed algorithm. The affiliated drivers were assigned recommendations based on the proposed algorithm (RLW) and a myopic business-standard algorithm alternatively and with three-hour strategy-flipping time interval. Over the day, cumulative results collected within the intervals with RLW recommendations are considered as the treatment group while the complementing outputs are associated to the control group. Note that the treatment drivers' dispatching mechanism was mainly structured according to the theoretical formulations given in this study, however, multiple minor variations were dictated by the business constraints to customize recommendations to corner cases such as new drivers, orders with very long waiting times, etc. 

\begin{figure}[h]
    \centering
    \includegraphics[scale=.22]{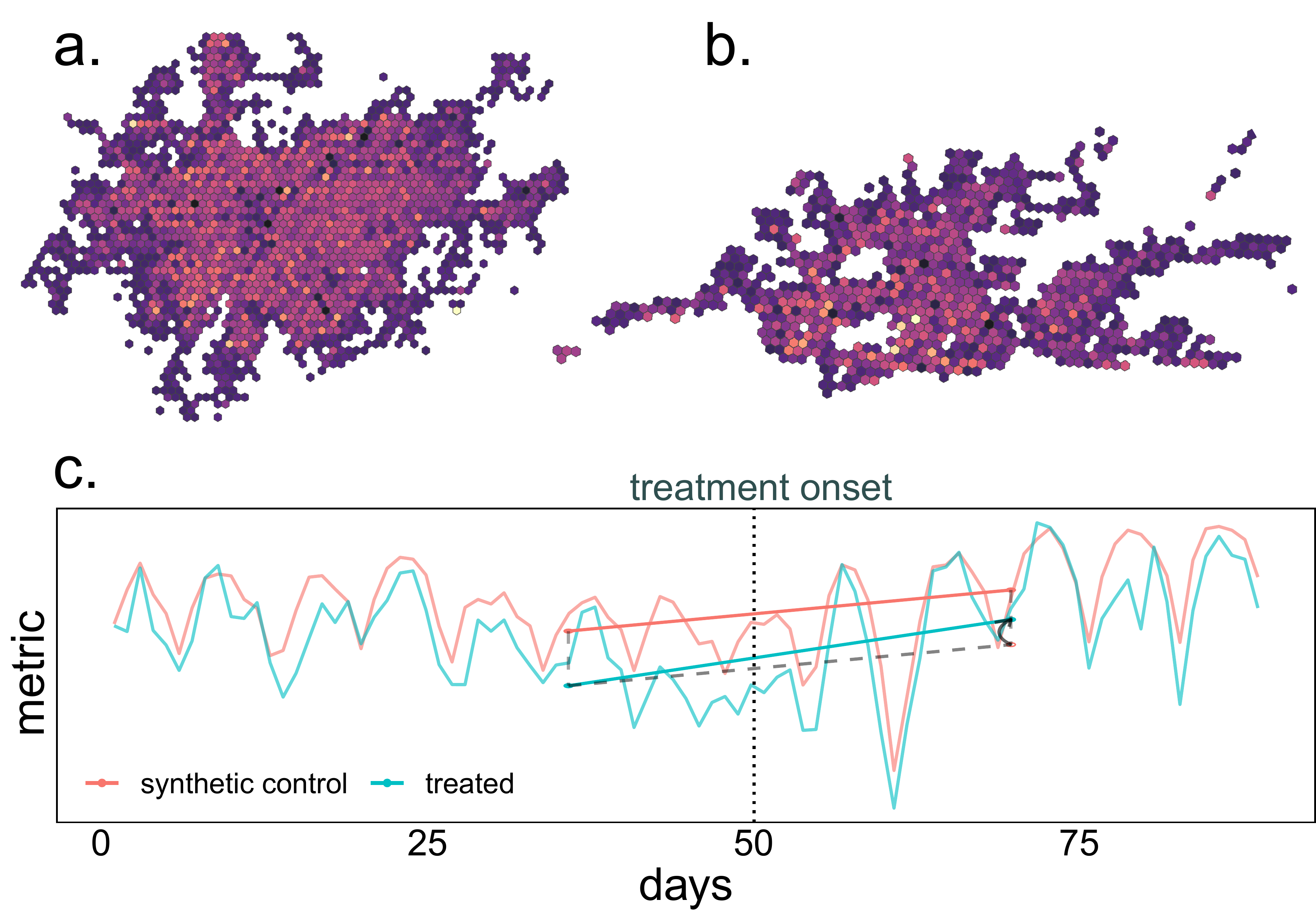}
    \caption{Deployment results. a-b: heatmap of spatial value functions in two cities at 10:00 AM of a weekday. The data is collected from A/B testing campaigns. The value heatmaps is learned through online training process. c: causal inference analysis on full-deployment case study. A major performance metric in the treated unit is compared against multiple control units before and after the onset of the treatment strategy. The analysis reveals a 5.350\% improvement after treatment.}
    \label{fig:heatmap}
\end{figure}

By processing the A/B testing experiment data, primary performance metrics are compared between control and treatment groups in Table \ref{tbl:AB}. The raw data is preprocessed to remove data points with partial and/or missing information and then, metrics are calculated based on the clean data. The values in the table denote the improvement of treatment compared to the control group in percentage. Overall, the treatment algorithm demonstrates notable improvements in terms of total driver income, CR, and AR. By comparing results for larger cities (Cities I, II, and IV) to the smaller areas, it is evident that the algorithm is more effective in the former group. We speculate that this observation stems from the fact that in smaller cities, supply and demand are sparse and therefore, myopic approach is quite efficient. By comparing the performance in A/B testing campaign with offline simulation, the direction of improvements are matching while the extents are different. We therefore argue that the simulation is fairly realistic and provide accurate insights and expectations. Note that the actual results might point to even higher improvement in the metrics if the experiment was performed in a low-variance supply/demand regime and was not bound to multiple business-related constraints that are highly tuned for the baseline model while not being adjusted with RLW.

In Figure \ref{fig:heatmap}-ab, the spatial distribution of value functions learned in two cities from A/B testing campaign are presented. Note that Figure \ref{fig:heatmap}-a represent the value distribution at 10:00 AM in City I while Figure \ref{fig:heatmap}-b presents the distribution in City II. The spatial value functions provide high-resolution preferences between available trip options for drivers that can increase the efficiency of the dispatching process in a farsighted manner (as opposed to the myopic approach). The value distribution reveals extremely proximal spatial coordinates with highly different values. Moreover, the map provides recommendations for the hot and cold areas for drivers in different times of the day. By comparing Figures \ref{fig:heatmap}-a and \ref{fig:heatmap}-b with reference to Table \ref{tbl:offline_res}, we hypothesize that a farsighted algorithm such as RLW shows better performance in cities with oval-shaped topology and uniformity of drivable areas. In City II (Figures \ref{fig:heatmap}-b), the topology is longitudinal and the drivable areas (presented by availability of hexbins) are patchy and distant. This can be correlated with the limited RLW improvement in City II. 
\vspace{-2mm}
\subsection{Full Deployment}


Based on successful results from A/B testing campaigns, the algorithm is now deployed in full-scale in one of the major international cities in DiDi's marketplace. Since the inception of the deployment at late December 2021, the deployment infrastructure has been responding to hundreds of thousands of daily requests by performing the RLW algorithm in real-time and returning recommendations to the affiliated drivers. Detailed overview of the infrastructure are presented in Figure \ref{fig:deploy} and Section \ref{sec:deployment} of the appendix. The response and engagement rates are highly affected by the contextual means such as holidays and seasonality and therefore, consistently fluctuating. In this deployment setting, the low latency module (Block B \& C from Figure \ref{fig:deploy}) computes dispatch recommendations in every two seconds for the batch of outstanding driver-order pairs. The communication rate by the low-latency module varies as a function of momentary demand and it is averaged to $\sim200$ calls per seconds. The maximum bipartite matching is a computationally expensive component of the full pipeline that performs Hungarian algorithm in every two seconds. The number of edges in the matching graph is dictated by the volume of outstanding order-driver pairs and can exceed 800 per second in peak hours. The low-latency module has been operating since the beginning of the deployment with no backlog or delay. 
\begin{table}[htp]
\centering
\caption{Summary of performance metrics from A/B testing campaign in four cities. Values indicate the ratio of metrics from the treatment group to the control group. Average metrics calculation is weighted based on the number of calls. The proposed method results in evident performance improvements in all metrics. Note that performance improvement is more significant in larger cities (City I, II, and IV).}
\label{tbl:AB}
\begin{tabular}{llllll}
\toprule
             & Income & CR       & AR     & \# samples\\ \hline
City I       & 1.977\%      & 0.260\% & 0.087\%& 4456166 \\
City II       & 1.255\%      & 0.680\% & 0.530\%& 6075712 \\
City IV      & 0.279\%      & 0.453\%  & 0.455\%& 1036612  \\
City V     & -0.739\%     & -0.822\% & -0.867\%& 469739  \\
City VI      & 0.145\%      & -1.102\% & -0.300\%& 189366  \\[0.04cm]
$\hat{X}_W$ &\textbf{1.342\%}      & \textbf{0.422\%} & \textbf{0.296\%}& -          \\ \bottomrule
\end{tabular}
\end{table}

In addition to the low-latency module, the infrastructure requires to communicate assignments to the high-throughput module in order to update the value function and other ever-changing parameters for different mechanisms (including reward smoothing, edge standardization, and learning rate adaptation) using the proposed algorithm.  In the current deployment, the rate of update for this module is set to 10 seconds. Therefore, all assignments within a 10-second time interval (equivalent to five rounds of dispatch) are passed the module for intermittent algorithmic updates. Block D (from Figure \ref{fig:deploy}) is receiving data with a rate of couple of thousands of calls in 10-second intervals. The module then updates the algorithmic tables and stores them in Block A (the data center) for further utilization in the proceeding rounds of dispatch.  

The deployment has been operational for a fairly short amount of time and therefore, a solid conclusion regarding the performance improvements may be possible in later times. In addition, as opposed the A/B testing campaign, a horizontal comparison between control and treatment group is impossible in a full deployment setting. Yet, we perform casual inference analysis to compare the performance of the deployment event historically with the dispatching performance prior to the deployment onset in the targeted city as well current performances in multiple other cities that are still following the traditional myopic approach. For this purpose, we use synthetic difference of differences method \cite{arkhangelsky2019synthetic} that is an interpretable extension of two widely used causal inference methods: synthetic control and difference of differences. In summary, these approaches aim to test the hypothesis that a treatment applied on a subset of units starting from a certain time has causal effects in certain temporal metrics. The time history of the designated metrics in the treated cities are regressed longitudinally (i.e., temporally) as well as vertically (i.e., by a weighted sum of controlled units) and then, regression is used to construct a counterfactual control trend for the metric in future times. The actual deviation in the designated metric with respect to the counterfactual (i.e., synthetic control) explains the causal effect of the treatment. 

By performing this analysis on the historical performance data from 37 cities (within the same geographical proximity), Figure \ref{fig:heatmap}-c is produced. The metric is a mixture of the introduced performance measures in this paper and is being anonymized and unscaled for business-related concerns.  The onset of the treatment - the first day of deployment - is marked with the vertical line. The synthetic control as well as the actual treated trendlines are plotted. The straight dotted line depicts the expected transition for the treatment trendline if the treatment was null or random. The curved arrow shows the extend of the actual improvement which cannot be explained by a random treatment and therefore, is attributed to the deployment. This analysis implies that the deployment has a clear positive impact - nearly 5.3\% - on the dispatching performance in the targeted city so far. In comparison to the findings from A/B testing, the rate of improvement is higher here. 
This is explained by some inherent issues in time flipping that are specifically required for the A/B testing \cite{tang2019deep} and usually underestimate the long-term effects. In addition, seasonality and nonstationary variations in supply and demand distributions can be effective in this variation. Regardless, both experimental settings strongly support the conclusion that the proposed model outperforms the state-of-the-practice. 
\vspace{-2mm}
\section{Conclusion}

The present work proposes a reinforcement learning framework for dispatching in ridehailing marketplace. The algorithm primarily builds upon online information from real-time trip requests and driver-order assignments by temporal difference (TD) updates and with no need for offline training on historical data. This feature enhances the scalability and generalization of the purposed algorithm compared to offline methods. Multiple algorithmic mechanisms are integrated with the online TD backbone to improve its efficacy. The expected update is replaced with the deterministic TD update to take into account the trip completion fate that is commonly determined after assignment. In addition, by integrating adaptive learning rates, state value updates are proportioned based on the frequency of state visitation as well as the extent of the update. To address high variance in immediate rewards (RL signals), a reward smoothing mechanism is incorporated. 

Regarding the maximum matching aspect of the dispatching algorithm, a customized utility function is proposed using which independent components are standardized and dynamically weighted. The utility function is then used for bipartite graph construction and matching. Finally, to control the rate of trip cancellation after driver assignment, a feedback control mechanism is designed and formulated as a multi-arm bandit that continually sets graph pruning thresholds based on real-time measures of the global performance metrics. An ensemble of these mechanisms is validated in three trials: (a) offline simulation based on real-world data, (b) full-scale A/B testing campaign, and (c) a full deployment in a major international market.  

By evaluation based on offline simulation, we showed that the proposed algorithm (RLW) outperforms state-of-the-art algorithms in majority of the tested cities. From A/B testing, it is confirmed that the deployed version of RLW is superior compared to the baseline in terms of total driver income, trip completion rate, and driver answer rate in larger cities. The improvement is marginal and occasionally negative in smaller cities which can be attributed to the sparsity of the city layout and demand. Finally, based on positive findings from former evaluations, the algorithm is deployed in full scale in a major international city in DiDi's marketplace. Historical trends for the primary performance metrics in the targeted city as well as $36$ other cities in its proximity are processed by synthetic difference of differences algorithm to identify treatment-related improvements. The analysis demonstrates a non-trivial improvement in the primary performance metric after the deployment. 


\bibliographystyle{ACM-Reference-Format}
\bibliography{ref}

%
\clearpage
\appendix
\setcounter{table}{0}
\renewcommand{\thetable}{A\arabic{table}}
\setcounter{figure}{0}
\renewcommand{\thefigure}{A\arabic{figure}}
\setcounter{algorithm}{0}
\renewcommand{\thealgorithm}{A\arabic{algorithm}}

\section{Deployment Detail}\label{sec:deployment}

The online deployment requires both low latency and high throughput in terms of communications and algorithmic operations. Once riders submit their trip requests, the backend service include them in the current batch of requests to be matched in the next round of order-driver (OD) assignments. This process is highly sensitive to unnecessary delays as riders may cancel their requests even before receiving the availability and pricing, thus low latency is a requirement. On the other hand, the RL-based algorithm requires continuous updates in the spatial value function as well as the algorithmic parameters such as adaptive learning rates, edge standardization, and reward smoothing variables. These updates can be performed in a pseudo-online fashion, yet the frequency and quantity of updates are high and therefore require high throughput capacity.

Based on these challenges, a backend infrastructure as schematically presented in Figure \ref{fig:deploy} is developed. In this setting, the data center (Block A) maintains a frequently updated record of necessary data required for the service, including log of requests, state value function, and algorithm parameters. The computational efforts are divided into two modules: low-latency and high-throughput module that are responsible for operations in the two segments in Algorithm \ref{alg:pipeline}. The former is responsible for quick preparation of the assignment problem by constructing the OD matrix (Block C) and solving the maximum bipartite matching using the Hungarian algorithm \cite{kuhn1955hungarian} (Block B) - \textit{order-driver assignment} in Algorithm \ref{alg:pipeline}. By these processes, the drivers dispatch is decided for the current batch of ODs and will be broadcast. The high-throughput module (Block D) on the other hand, does not need to be rapidly responsive. Instead, in this module, the algorithmic updates take place in every K second intervals (K is a hyperparameter). This module is specialized in processing large amount of data (OD pairs assigned in the last K seconds) with fairly high frequency. In Block D, the second segment of the algorithm - \textit{state value updating} from Algorithm \ref{alg:pipeline} - is performed.  

Some of the algorithmic mechanisms introduced in Sections \ref{sec:MDP} and \ref{sec:matching} including adaptive learning rates, reward smoothing, and edge standardization are ideally grid-level or order-level updates and can take place simultaneously along with the request reception. In the full-scale, however, this is found impractical since it requires extra computational overhead. To address that, we revised the introduced rules to the batch-mode versions and integrated them with Block D - see lines \ref{line:rew_smooth} or \ref{line:std_update1} in Algorithm \ref{alg:pipeline} as examples. This approach was evaluated offline against the ideal approach and the performance reduction was quite negligible. From multiple algorithmic mechanisms introduced in the present work, the graph pruning using LM-UCB could not be deployed in large-scale due to more involved deployment challenges. In particular, in the current infrastructure, there is no simple way to query a real-time or mildly-delayed global feedback for the algorithm performance metrics. 

\vspace{-2mm}
\section{Practical Notes} 

\begin{figure}
    \centering
    \includegraphics[scale=0.73]{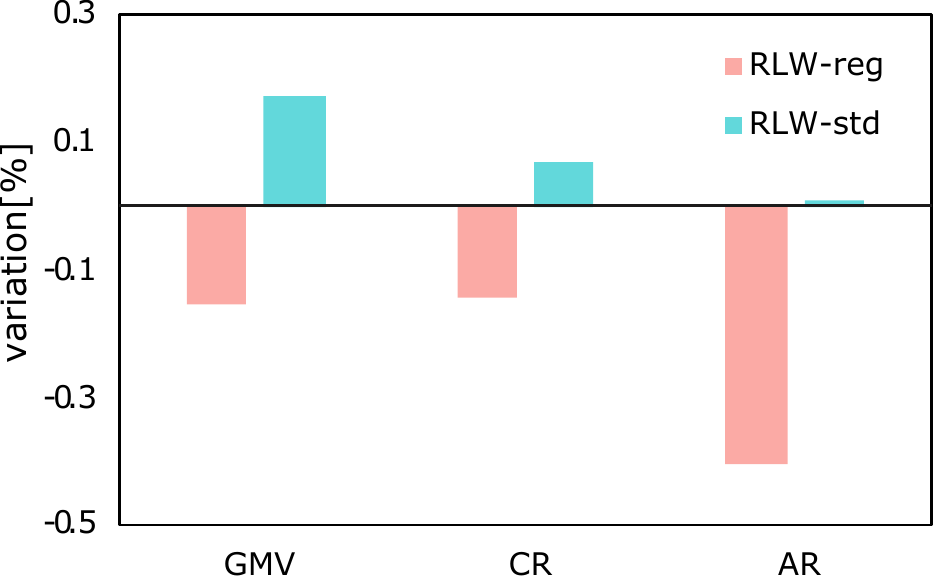}
    \caption{RLW is insensitive to price scale: the variations in major metrics caused by price scaling are compared in a fixed day and city between two models: RLW with regular graph edges (RLW-reg) vs. RLW with edge standardization (RLW-std). Plot shows the mean of variations in the two cases - prices scaled by factor 0.5 and 2 - with respect to the unscaled case. }
    \label{fig:WF}
\end{figure}

In this study, we performed the algorithm in three levels and confirmed a consistent improvement in the performance metrics. By combining the observations, some practical insights can be inferred. We define the success rate (SR) as a ratio between the number of completed trips and the total number of dispatches. Ideally, a dispatching algorithm is perfect when both AR and SR are close to one - all requests are answered and all answered trips will be completed. 

\begin{figure*}[!htp]
    \centering
    \includegraphics[scale=.65]{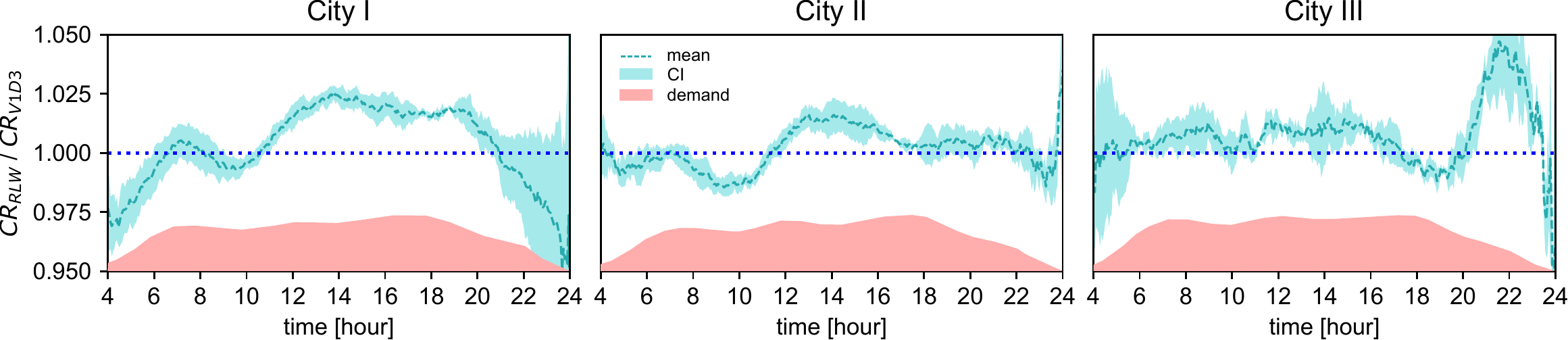}
    \caption{CR ratio between RLW and V1D3 over the day in the three cities. Demand temporal patterns are shown by the shades. At any given time, CR is calculated based on the dispatching logs of the latest five minutes. For each city, the graphs are plotted for three days. The CR ratio between two models show that overall, RLW model starts from a inferior performance point initially, however, over the day, it outperforms V1D3 and its performance remains superior within the high-demand time periods.}
    \label{fig:cr-trend}
\end{figure*}

By monitoring the performance of state-of-the-art models, we argue that there are two apposing factors that affect the overall performance. The first factor is associated with improving AR by maximizing the collocation between drivers and demand. CVNet-B is an ultimate case in this aspect since it is trained by a large corpus of historical data. This can be confirmed by large AR improvements in CVNet-B from Table \ref{tbl:offline_res}. However, offline training is prone to major disadvantages such as lack of adaptability to distribution shifts. In the RLW, we incorporate multiple mechanisms including the expected update, adaptive learning rate, and reward smoothing to improve AR. On the other hand, when AR is improved, the quantity of assignment increases and consequently, the rate of false positives (assignments that will be eventually canceled) rises - and SR declines. Therefore, a second force should also be considered: graph pruning and edge penalty mechanisms for increasing SR. We introduce mechanisms such as LM-UCB and graph edge standardization to address this factor. When two factors interact efficiently, a more efficient trade-off between AR and SR can be achieved. 


\subsection{Limited-Memory UCB}

LM-UCB is used to control the rate of cancellation in a feedback-based scheme. The algorithm is presented in Algorithm \ref{alg:USB}.
\begin{algorithm}
\caption{Limited-Memory UCB (LM-UCB)}\label{alg:USB}
\begin{algorithmic}
\Require $c, \alpha, \gamma, Arms$
\State $Q_i \gets 0$ \Comment{for $i \in$ Arms}
\State $N_i \gets 0$ \Comment{for $i \in$ Arms}
\State $th \gets RandomChoice\{Arms\}$
\State $n \gets 0$
\For{$t = 0:T$} \Comment{$t$ in seconds}
\State Maximum Bipartite Matching($th$)
\State Value Updating
\If{$mod(t,60)=0$}
    \State $n \gets \gamma n + 1$
    \State $N_{a} \gets \gamma N_{a}: \forall a \in Arms$
    \State $q \gets CR + 0.1 \times AR$ \Comment{tentative function}
    \State $Q_{th} \gets \alpha Q_{th} + (1 -\alpha) q$
    \State $N_{th} \gets N_{th}+1$
    \State $th \gets argmax\{Q_a + c \sqrt{\frac{\log{n}}{N_a}}: \forall a \in Arms\}$
\EndIf
\EndFor
\end{algorithmic}
\end{algorithm}
In this algorithm, $\alpha$ and $\gamma$ are hyperparameters that determine the memory length of the UCB mechanism, $Arms$ is a set of discrete probability values (here $Arms = linspace(0,0.3,41)$). $c$ is the exploration promoting factor. In this algorithm, a tentative performance metric (here $CR + 0.1AR$) is evaluated every minute, and then, the expected value of the current arm $th$ is updated based on this recent feedback using an exponentially moving average formula. In addition, the arm exploitation frequency $N_{th}$ and total counter $n$ are discounted to induce a limited memory. In this problem, since the performance metrics are non-stationary, a short-memory mechanism is effective to adapt more rapidly to these variations. Compared to the vanilla version of UCB, the proposed approach is found more effective. One may argue that the existing non-stationarity encourages the use of contextual bandit solutions such as linUCB \cite{li2010contextual}. In fact, by conducting comparative evaluations, we confirmed that the limited-memory approach outperforms the contextual solutions as well. We hypothesize that the linUCB model requires longer training duration (multiple days) and significantly more updates (compared to one update per minute in LM-UCB) in order to converge to a stable policy. 

\subsection{Standardization Effect}
Graph edge standardization enables to mix multiple factors in the maximum matching problem in a controlled and scale-invariant setting. Unlike the predecessor of RLW, the components of the edge weights are bounded values that are mixed by user-defined factors. The benefits of this framework are twofold. First, in online training, the weight factors can adjust the contribution of the value function and penalty terms dynamically. Second, when incorporating penalty terms, the associated weight factors are insensitive to units and scales, improving transferability of simulation hyperparameters to real-world applications as well as new domains. To examine the latter characteristic, two variants of RLW model are compared: (1) RLW-reg in which the edges are defined as the sum of unscaled and unfactored immediate reward, residual value, and penalty term and (2) RLW-std in which edges are comprised of standardized and weighted components. Models are simulated against real data from City II in three scenarios: considering (a) unscaled trip prices, (b) doubled prices, and (c) halved prices for value updating as well as the maximum matching problem. Results are summarized in Figure \ref{fig:WF}, in which the total variations in performance metrics from the two scaled scenarios - cases (b) and (c) - with respect to the unscaled scenario are compared. The result shows that overall, the range of variations in scaled scenarios is significantly smaller in RLW-std model. In addition, RLW-reg model shows negative variations in scaled cases since its penalty term is tuned with reference to the unscaled price units and is incompatible to the arbitrarily scaled scenarios.This confirms that standardization minimizes the variation with respect to price scaling (e.g., currency, unit systems, etc) while RLW-reg shows a consistent drop in performance with price scale changes.

In Figure \ref{fig:cr-trend}, a brief comparison between V1D3 and RLW is presented. The comparison is performed based on CR trend variations over a full day in three cities and three days. The dotted curve depicts the mean of the three daily trials and the standard deviation is presented with the surrounding envelope. The shady area on the bottom also depicts a schematic of the temporal demand distribution. The plots demonstrates that in most cases, RLW yields a higher CR when the demand is high in comparison with V1D3. In addition, V1D3 performs better at the beginning and end of the days. RLW outperforms V1D3 in the peak hours and for the majority of the operational time. We speculate that the initial underperformance can be attributed to the use of smoothed RL signals and expected updates that can slow down the process of learning, but increase its robustness. Note that practically, V1D3 is a special case of a the class of RLW models in which introduced mechanisms are ineffective. The comparisons in Figure \ref{fig:cr-trend} and Table \ref{tbl:offline_res} imply that RLW provides a suite of tunable knobs such as reward smoothing coefficient, standardization coefficients, and LM-UCB that can improve the performance of the vanilla V1D3 algorithm in widely dissimilar supply-demand distributions.

\end{document}